\documentclass[runningheads]{llncs}

\usepackage{eccv}

\usepackage{eccvabbrv}

\usepackage{graphicx}
\usepackage{booktabs}
\usepackage{array}
\usepackage{caption}
\usepackage{multirow}
\usepackage{ragged2e}
\usepackage{tabularray}
\usepackage{wrapfig}
\RequirePackage[dvipsnames]{xcolor}
\RequirePackage{amssymb}
\RequirePackage{pifont}
\RequirePackage{times}    
\RequirePackage{xspace}
\RequirePackage{amsmath}
\usepackage{pdfpages}

\usepackage[accsupp]{axessibility}  

\usepackage{hyperref}

\usepackage{orcidlink}

\begin{document}

\title{MedCAGD: Context-Aware Gated Decoder for Efficient Medical Image Segmentation} 

\titlerunning{MedCAGD}

\author{
Saad Wazir\inst{1}\orcidlink{0000-0001-9260-1636} \and
Patrick Dominique Vibild\inst{2}\orcidlink{0000-0001-9437-0149} \and
Dinh Phu Tran\inst{1}\orcidlink{0009-0003-4254-2158} \and
Seongah Kim\inst{1}\orcidlink{0009-0005-0414-4399} \and
Daeyoung Kim\inst{1}\orcidlink{0000-0002-7960-5955}
}

\authorrunning{S. Wazir et al.}

\institute{
School of Computing, Korea Advanced Institute of Science and Technology (KAIST), Daejeon, Republic of Korea \\
\email{\{saad.wazir,phutx2000,kimsa0322,kimd\}@kaist.ac.kr} \and
Department of Energy, Aalborg University, Aalborg, Denmark \\
\email{padovi@energy.aau.dk}
}

\maketitle

\begingroup
\renewcommand{\thefootnote}{}
\footnotetext{Accepted at the European Conference on Computer Vision (ECCV 2026).}
\addtocounter{footnote}{-1}
\endgroup

\begin{abstract}
Medical image segmentation relies on the ability of encoder-decoder architectures to translate rich feature representations into accurate pixel-level predictions under challenging conditions such as low contrast, structural ambiguity, and scale variability. While recent advances in large-scale pretraining and transformer-based encoders have substantially improved feature extraction, segmentation accuracy remains constrained by decoder design, particularly in terms of cross-scale alignment, contextual integration, and boundary preservation. In this work, we revisit medical image segmentation from a decoder-centric perspective and propose a context-aware gated decoder that systematically regulates feature fusion and contextual aggregation throughout the decoding process. The proposed decoder integrates lightweight multi-scale channel recalibration, gated skip fusion with spatial competition and a global context aggregation mechanism that injects encoder-wide information into intermediate decoding stages. This design enables effective translation of strong pretrained encoder representations into spatially consistent predictions. Extensive experiments across 11 medical image segmentation benchmarks validate the effectiveness and  demonstrate that the proposed approach consistently outperforms strong baselines while remaining computationally practical. Code: \href{https://github.com/saadwazir/MedCAGD}{https://github.com/saadwazir/MedCAGD}
\end{abstract}

\section{Introduction}
\label{sec:Introduction}
Medical image segmentation is fundamental for quantitative analysis, diagnosis, treatment planning, and clinical assessment. Tasks such as organ delineation, lesion localization, tumor boundary extraction, and cellular segmentation require pixel-level precision under challenging conditions. To address these challenges, encoder–decoder architectures, particularly U-Net \cite{u-net} based models, have become the dominant paradigm in medical image segmentation. Within this framework, performance is increasingly governed by encoder–decoder design, skip-connection formulation, feature fusion, and, more prominently, improvements in encoder capacity \cite{deepchallenge, mist, pvt-cascade-skin}. Although attention mechanisms have evolved from local CNN-based modules to non-local and transformer-based formulations for long-range context modeling \cite{attention-mech, advantages-transformer}, their computational cost and design limitations leave the integration of global context during decoding unresolved. Consequently, many segmentation errors arise from suboptimal decoding and cross-scale alignment rather than insufficient feature extraction \cite{deep-method-survey,emcad}.

Recently, foundation model approaches have demonstrated strong cross-domain generalization in vision tasks. In segmentation, SAM \cite{sam} has introduced a promptable, generalist paradigm, inspiring SAM-derived medical variants \cite{autosam, customizedsam, sam3d, self-promptsam} that demonstrate strong generalization. However, even medically adapted versions require substantial labeled data, modality specific supervision, and significant computational resources to approach the performance of specialist medical segmentation models \cite{medsam}.

In parallel, advances in large-scale pretraining have strengthened encoder representations \cite{convnext1, convnext2, pvt1, pvt2, maxvit}, leading modern segmentation frameworks to adopt powerful pretrained encoders \cite{mist, cascaded-Aattention, pranet}, yet accuracy remains dependent on effective decoding into spatially consistent predictions, with boundary errors and fragmentation often stemming from semantic misalignment rather than weak representations \cite{emcad, mcads-decoder}. Taken together, these observations indicate that continued improvements in medical image segmentation accuracy increasingly hinge on decoder design rather than encoder capacity alone. In this context, this work explores decoder design as a complementary and computationally efficient approach for improving segmentation accuracy. Designing such decoders remains challenging, as they must balance contextual integration with spatial precision without incurring excessive computational cost. We argue that meaningful performance gains can be achieved through a principled decoder centric framework that systematically regulates contextual aggregation across decoding stages, enabling more faithful translation of strong encoder representations into accurate pixel-level predictions without increasing encoder complexity or relying on task specific fine tuning. Our main contributions are:
\begin{itemize}
    \item
    \textbf{MedCAGD: Context-Aware Gated Decoder Architecture.}
    We propose a decoder centric segmentation framework that systematically regulates feature transformation during decoding. The architecture integrates Bottleneck with Global Context Injection, Spatially Competitive Attention Gate based skip regulation, Multi-level Context aggregation, and stage wise refinement, positioning decoder design as the primary factor governing accurate pixel-level prediction.
    \item
    \textbf{Structured Context Regulated Decoder Components.}
    We introduce a unified set of modules that directly correspond to the methodological components of MedCAGD:
    \textbf{(i) Efficient Channel Attention Block with multi-scale Pooling,} which performs context sensitive channel recalibration using multi-scale descriptors and normalized channel competition.
    \textbf{(ii) Spatially Competitive Attention Gate,} which formulates skip fusion as normalized multiplicative encoder-decoder agreement combined with global modulation and multi-scale spatial competition.
    \textbf{(iii) Multi-level Context Aggregation with Residual Attention,} which injects globally coherent multi-level encoder semantics into intermediate decoder stages to mitigate cross scale semantic misalignment.
    \textbf{(iv) Refinement Block} with local refinement and channel recalibration, which strengthens local reconstruction and stabilizes feature propagation across decoding stages.
    \item
    \textbf{Encoder agnostic and computationally efficient design with strong empirical validation.}
    The proposed MedCAGD remains fully encoder-agnostic through Universal Feature Projection, enabling broad compatibility with PyTorch timm encoders, while maintaining practical complexity of \textbf{30.60 M parameters} and \textbf{5.0 GFLOPs}. Extensive experiments across 11 heterogeneous medical image segmentation benchmarks demonstrate consistent improvements over strong CNN, Transformer, Mamba, SAM, and recent decoder centric baselines.
\end{itemize}

\section{Related Work}
\label{sec:Related Work}
\textbf{CNNs} have been the cornerstone of medical image segmentation, most notably U-Net \cite{u-net}, which became dominant by combining hierarchical features with skip connections to recover fine spatial detail. Building on this design, a wide range of U-Net variants \cite{unet++, unet3+, u2-net, ren-unet, histoseg, histoseg++} emerged. These works introduced dense skip connections, nested U-Net designs, and multi-scale aggregation for improved context and boundaries, while nnU-Net \cite{nnU-Net} highlighted the role of systematic pipeline optimization. However, CNNs still rely on local operations, limiting long-range dependency modeling. \textbf{Attention mechanisms} \cite{attentionu-net, scau-net, raunet} partially mitigate this by enhancing features via channel, spatial, and residual attention, but mainly recalibrate features without modeling global interactions.

\textbf{Transformer} based architectures address the limitation of long-range dependency modeling by introducing self-attention. TransUNet \cite{transunet}  pioneered the integration of Vision Transformers with convolutional decoders. Subsequent architectures such as Swin-Unet \cite{swin-unet} adopted hierarchical Transformer designs with shifted window attention to improve computational efficiency. Some architectures such as UNeXt \cite{unext} replace self-attention with convolutional MLP-based designs to reduce computational overhead, while task-specific models such as PraNet \cite{pranet} introduce structurally motivated attention mechanisms to enhance boundary cues without relying on full global attention. However, Transformers suffer from quadratic computational and memory complexity, limiting scalability.

\textbf{Mamba} \cite{mamba} addresses the quadratic computational and memory inefficiency of Transformers by replacing explicit attention with linear time state space modeling. Several recent works have explored Mamba-based architectures for medical image segmentation. VM-UNet \cite{vm-unet} introduces Vision Mamba blocks into a U-Net style architecture to enhance long-range spatial dependency modeling while maintaining linear computational complexity. U-Mamba \cite{u-mamba} further integrates Mamba blocks into CNN encoders within the nnU-Net framework, combining local convolutional feature extraction with state space modeling to improve global context representation. Swin-UMamba \cite{swin-umamba} extends this by incorporating hierarchical representations and ImageNet pretrained Mamba-based encoders. Existing Mamba-based segmentation methods primarily emphasize encoder representations and typically operate with a fixed state size, which may limit performance scalability across tasks of varying complexity.

\textbf{Decoder} design has been advanced through multi-scale context aggregation \cite{deeplayer}, dense \cite{unet++} or full scale skip connections \cite{unet3+}, deep supervision \cite{mu-net, dseu-net}, efficient spatial reconstruction modules \cite{mist}, dual decoder architectures \cite{ddanet}, and the integration of transformer blocks \cite{swin-unet}. UCTransNet \cite{uctransnet} replaces fixed skip connections with learnable semantics aware fusion to better align multi-scale features while preserving spatial detail, while PolypPVT \cite{polyppvt} embed CBAM \cite{cbam} within the decoding stage for enhanced feature refinement. MCADS \cite{mcads-decoder} follows a complementary direction inspired by \cite{raunet, understandingconv}, combining residual linear attention with depth to space based upsampling to preserve fine structural details during resolution recovery, achieving higher accuracy at the expense of efficiency. More recently, EMCAD \cite{emcad} introduces a convolutional decoder that integrates multi-stage hybrid transformer encoder features using modified and enhanced attention mechanisms following \cite{squeeze, attentionu-net, cbam}, leading to strong performance in medical image segmentation. Despite recent progress, segmentation performance remains fundamentally constrained by decoder design. In particular, challenges in cross-scale alignment, boundary refinement, and long-range context translation continue to persist across the previously discussed architectures. Although these methods introduce increasingly sophisticated attention, upsampling, and feature fusion strategies, they often prioritize stronger pretrained encoders while relying on decoding mechanisms that inadequately preserve fine spatial detail and global semantic consistency. Consequently, they achieve only modest improvements and inconsistent performance across tasks of varying anatomical complexity and structural variability. Collectively, these observations indicate that the primary bottleneck lies in decoder formulation rather than encoder capacity alone, motivating the exploration of more principled and context-aware decoder designs for medical image segmentation.

Foundation models, particularly \textbf{Segment Anything Model (SAM)} \cite{sam} has demonstrated strong generalization across diverse image segmentation tasks through prompt driven interaction, enabling flexible mask generation. In the medical imaging domain, several adaptations such as AutoSAM \cite{autosam}, Medical SAM3 \cite{medicalsam3}, SAMed \cite{customizedsam}, SAM3D \cite{sam3d} and Self-Prompt-SAM \cite{self-promptsam}  have explored fine tuning strategies, adapter based training, and learned prompting mechanisms to better align SAM with domain specific structures. While these approaches improve robustness, they typically rely on explicit prompting, large curated datasets for adaptation, and substantial computational resources. Empirical studies further indicate that SAM based methods often under-perform specialist architectures on fixed task, particularly for datasets characterized by subtle boundaries or fine grained anatomical structures, such as fundus imaging \cite{medsam}. Although not the primary focus of this study, SAM based methods are included to enrich the analysis and provide a broader contextual understanding.

\begin{figure*}
\centerline{\includegraphics[width=1\textwidth]{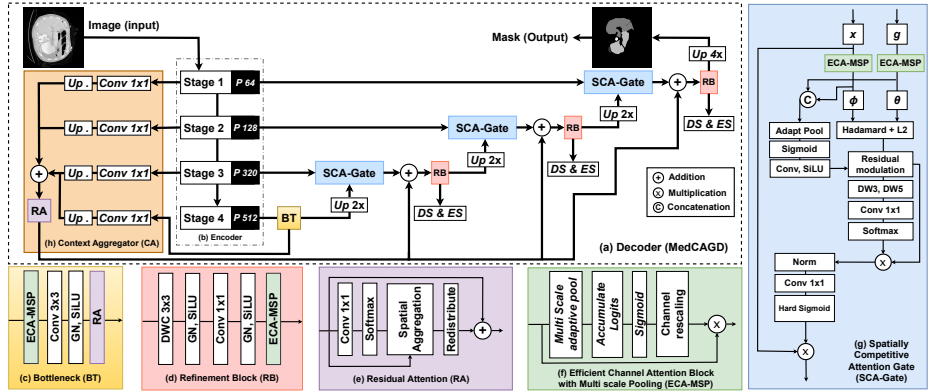}}
\caption{Overview of \textbf{(a)} MedCAGD, the proposed decoder architecture. \textbf{(b)} Multi-scale encoder features are projected into a unified decoder feature space. \textbf{(c)} Bottleneck (BT) initializes decoding by refining the deepest encoder feature using \textbf{(f)} Efficient Channel Attention with Multi-scale Pooling (ECA-MSP) for adaptive channel recalibration and \textbf{(e)} Residual Attention (RA) for global context integration. \textbf{(g)} Spatially Competitive Attention Gate (SCA-Gate) selectively regulates encoder skip features before fusion with decoder features. \textbf{(h)} Context Aggregator (CA) injects globally aggregated multi-scale semantics into each decoding stage. \textbf{(d)} Refinement Block (RB) enhances fused decoder features through efficient local refinement and channel recalibration. Deep Supervision (DS) and Edge Supervision (ES) provide auxiliary supervision during training.
}
\label{fig:arch}
\end{figure*}

\section{Methodology}
In this section, we first present the overall encoder–decoder architecture and explain how its components are integrated to regulate feature flow during decoding. We then describe the fundamental modules that form the foundation of the proposed method, along with a brief introduction to the encoder. Finally, we introduce the training objective. The complete pipeline of the proposed approach is depicted in Fig. \ref{fig:arch}.

\subsection{Overall Decoder Architecture and Component Integration}
Multi-scale hierarchical features are first extracted by the encoder and projected into fixed dimensional representations. Decoding begins from the bottleneck output and proceeds through a sequence of decoder blocks with progressively increasing spatial resolution. At each stage, the current decoder feature is first upsampled and then fused with the corresponding encoder feature through the Spatially Competitive Attention Gate, enabling selective and context-aware regulation of encoder features prior to concatenation. In parallel, Multi-level Context Aggregation operates on the projected encoder features, and the resulting context representation is added residually to the decoder feature before refinement. This ensures that each stage is guided by globally aggregated multi-scale semantics while preserving stage-specific reconstruction. The updated feature is then passed through the Refinement Block for convolutional enhancement and channel recalibration. This sequence of upsampling, gated skip fusion, context aggregation, and refinement is repeated across decoding stages. At the final stage, the full resolution decoder feature is forwarded to the segmentation head to produce the primary prediction. Intermediate decoder features are additionally connected to auxiliary segmentation and edge prediction heads to enable deep supervision during training.

\subsection{Encoder and Universal Feature Projection}
We employ an ImageNet pretrained PVTv2-B2 as the encoder due to its hierarchical transformer design, which provides multi-level features well aligned with our decoder. It achieves a strong balance between accuracy and efficiency, as validated in Sec. \ref{sec:ablback}. Furthermore, its adoption by several SOTA decoder centric methods ensures fair comparison. As shown in Fig. \ref{fig:arch} (b), given multi-scale encoder feature maps $\{c_i\}_{i=1}^{4}$, each encoder feature $c_i$ is aligned to the predefined decoder channel dimension using a learnable $1 \times 1$ convolutional projection $\mathcal{P}_i(\cdot)$, such that $p_i = \mathcal{P}_i(c_i)$, where the decoder channels are fixed to $64$, $128$, $320$, and $512$ across stages. This projection ensures consistent decoder dimensionality while preserving the multi-scale hierarchy of the encoder.

\subsection{Residual Attention (RA)}
To model global spatial dependencies during decoding, we employ a lightweight non-local attention mechanism embedded within a residual formulation. As shown in Fig. \ref{fig:arch} (e), given an input feature map $X$, a spatial importance distribution is first computed using a pointwise projection followed by softmax normalization. This distribution is used to aggregate long-range spatial responses into a global context descriptor. The aggregated vector is then transformed through a lightweight channel mixing function with intermediate dimensionality reduction and reinjected into the feature stream via residual addition, yielding
\begin{equation}
Y
=
X
+
\mathcal{P}_2
\Big(
\delta
\big(
\mathcal{P}_1
(
\sum_{i=1}^{HW}
\mathrm{Softmax}
(
\mathcal{P}_0(X)
)_i
\, X_i
)
\big)
\Big).
\end{equation}
Here, $\mathcal{P}_0(\cdot)$, $\mathcal{P}_1(\cdot)$, and $\mathcal{P}_2(\cdot)$ denote learnable pointwise convolutional projections, and $\delta(\cdot)$ denotes a nonlinear activation. The residual formulation preserves local structure while enabling efficient global context integration, as reported in Sec. \ref{sec:ablcomponent}, where enabling RA improves performance.

\subsection{Efficient Channel Attention Block with Multi-scale Pooling (ECA-MSP)}
To adaptively recalibrate channel responses based on contextual relevance, we employ an Efficient Channel Attention block extended with multi-scale pooling. As shown in Fig.~\ref{fig:arch}(f), given an input feature map $X$, channel descriptors are extracted at multiple contextual granularities using adaptive average pooling. Here, the pooling scale refers to the target spatial resolution of adaptive average pooling used to compute channel statistics, while the pooling operation itself remains average pooling. For a set of pooling scales $\mathcal{S} = \{1, 2, 4\}$, multi-scale channel descriptors are independently transformed through a one-dimensional convolution that models local cross channel interaction without dimensionality reduction. The resulting responses are aggregated across scales and converted into channel attention weights. The overall operation is expressed as
\begin{equation}
X' = X \odot \sigma\!\left(A_{\text{ms}}(X)\right),
\quad
A_{\text{ms}}(X)
=
\frac{1}{|\mathcal{S}|}
\sum_{s \in \mathcal{S}}
\psi\!\left(\mathrm{AdaptiveAvgPool}_s(X)\right).
\end{equation}
where $\mathrm{AdaptiveAvgPool}_s(\cdot)$ denotes adaptive average pooling to spatial size $s \times s$ followed by spatial aggregation to obtain channel descriptors, $\psi(\cdot)$ denotes local cross channel interaction implemented via one dimensional convolution, and $\sigma(\cdot)$ denotes the sigmoid activation. Unlike SE-Net \cite{squeeze}, ECA-Net \cite{eca-net}, and EMCAD \cite{emcad}, which rely on single scale global descriptors or bottleneck based dual pooling, the proposed formulation leverages multi-scale pooling to capture complementary contextual information, ranging from global semantic statistics to coarse localized cues. Sec. \ref{sec:ablcomponent} and \ref{sec:ablgate} provides empirical evidence of the effectiveness of ECA-MSP.

\subsection{Bottleneck (BT) with Global Context Injection}
At the deepest stage of the network, a bottleneck module refines the highest level encoder feature and injects global context before decoding begins as shown in Fig. \ref{fig:arch} (c). Let $F_4$ denote the projected deepest encoder feature. Channel responses are first recalibrated using the ECA-MSP $\mathcal{E}(\cdot)$, followed by convolutional refinement $\rho(\cdot)$, and finally global context injection through the RA operator $\mathcal{R}(\cdot)$. The overall bottleneck transformation is expressed as $B = \mathcal{R}\!\left( \rho\!\left( \mathcal{E}(F_4) \right) \right).$
This formulation enables the decoder to start from a context-aware semantic representation while preserving the structural properties of the refined feature map. Further validated by ablation study in Sec. \ref{sec:ablcomponent}.

\subsection{Spatially Competitive Attention Gate (SCA-Gate)}
Recent studies \cite{uctransnet, udtransnet} indicate that skip connections in encoder-decoder architectures are not universally beneficial, since indiscriminate feature propagation can introduce semantically incompatible information due to the encoder-decoder semantic gap. Following this motivation, we formulate skip connections as learnable and selective feature regulation mechanisms rather than passive information pathways, as shown in Fig.~\ref{fig:arch} (g). Let $g$ denote the decoder feature at a given stage and $x$ the corresponding encoder skip feature. Both features are first recalibrated using ECA-MSP $\mathcal{E}(\cdot)$ and projected into a shared latent space via lightweight transformations $\theta(\cdot)$ and $\phi(\cdot)$. Their interaction is modeled as $f = \theta(\mathcal{E}(g)) \odot \phi(\mathcal{E}(x))$. The gated skip feature is then defined as
\begin{equation}
\begin{aligned}
x' &= x \odot \sigma\!\left(\mathcal{H}(f,g,x)\right), \\
\mathcal{H}(f,g,x)
&= f \odot \big(1 + \mathcal{G}(g,x)\big)
   \odot \big(1 + \mathcal{S}(f)\big).
\end{aligned}
\end{equation}
Here, $\mathcal{G}(\cdot)$ denotes global channel modulation derived from the joint encoder-decoder representation, while $\mathcal{S}(\cdot)$ represents multi-scale spatial competition. In practice, $\mathcal{S}(f)$ is implemented using parallel depthwise convolutions with kernel sizes $3$ and $5$, namely $\mathcal{D}_3(f)$ and $\mathcal{D}_5(f)$, whose aggregated responses are normalized via temperature-controlled softmax. The resulting attention mask $\sigma(\mathcal{H}(f,g,x))$ is multiplicatively applied to the skip feature $x$. The function $\sigma(\cdot)$ denotes a bounded activation for adaptive skip regulation. Unlike Attention U-Net \cite{attentionu-net} and EMCAD \cite{emcad}, which rely on additive fusion followed by sigmoid masking, the proposed formulation models skip selection as normalized multiplicative agreement combined with global modulation and spatial competition. The effectiveness of SCA-Gate is further validated by comprehensive ablation studies in Sec. \ref{sec:ablcomponent} and \ref{sec:ablgate}.

\subsection{Context Aggregator (CA)}
While skip connections transfer information between corresponding encoder and decoder stages, effective decoding also requires global awareness across multiple semantic scales. To this end, we introduce a multi-level context aggregation module, as shown in Fig.~\ref{fig:arch}(h), which integrates features from different levels and injects globally consistent contextual information into each decoding stage  as supported by the ablation study in Sec.~\ref{sec:ablcomponent}. Let $\{F_k\}_{k=1}^{K}$ denote feature maps from multiple encoder stages. Each feature is projected into a unified channel space and spatially aligned to the target decoder resolution using learnable pointwise projections $\mathcal{P}_k(\cdot)$ with interpolation. The aligned features are averaged and refined through the RA operator, producing $F_{\mathrm{ctx}} = \mathcal{R}\!\left( \frac{1}{K} \sum_{k=1}^{K} \mathcal{P}_k(F_k) \right)$. The resulting representation aggregates globally coherent multi-scale semantics and is added residually to the decoder feature, providing stage independent global guidance that complements context-aware gated skip fusion.

\subsection{Refinement Block (RB)}
The Refinement Block, shown in Fig.~\ref{fig:arch}(d), enhances decoder feature propagation through sequential local refinement and adaptive channel recalibration, as validated by the ablation study in Sec. \ref{sec:ablcomponent}. Given an input feature, it is first processed by a depthwise convolution for spatial filtering, followed by a pointwise convolution for channel mixing, each combined with Group Normalization and SiLU activation. This design enhances spatial consistency while maintaining computational efficiency during local refinement. The refined features are subsequently recalibrated using ECA-MSP.

\subsection{Segmentation Outputs and Training Objective}
The decoder produces the final segmentation along with auxiliary segmentation and edge predictions for supervision. The final feature generates the primary logit, while intermediate features are independently projected and upsampled to the input resolution for deep supervision. Let $\hat{Y}$ denote the final segmentation logit and $\{\hat{Y}_i\}_{i=1}^{3}$ the auxiliary segmentation logits. Deep supervision is applied by optimizing a weighted and normalized sum of losses over these predictions to promote consistent optimization across decoding depths. In parallel, auxiliary edge logits $\{\hat{E}_i\}_{i=1}^{3}$ are generated from intermediate decoder features and supervised using binary edge targets derived from the ground truth masks, encouraging boundary aware decoding. All segmentation and edge predictions are optimized using the binary cross entropy (BCE) loss \cite{losssurvey}. The overall training objective combines the main segmentation loss with deep supervision and edge supervision losses using normalized weights. Ablation results in Fig. \ref{fig:abl4} demonstrate that enabling both deep supervision (DS) and edge supervision (ES) consistently improves Dice and HD95 across six segmentation benchmarks, highlighting their complementary roles in dense prediction and contour refinement. During inference, only the final segmentation prediction is retained.

\begin{table*}[]
\centering
\caption{Comprehensive performance comparison across 9 medical image segmentation benchmarks. Average Dice scores $\uparrow$ are reported. Bold and underline denote the best and second best results, respectively. All methods were reproduced and averaged over five runs, with fine tuning applied to SOTA models for fair comparison. Results marked with * are reported from the papers. “–” indicates unavailable results.}
\label{tab:tab1}
\resizebox{1\columnwidth}{!}{%
\begin{tabular}{@{}l|c|c|cc|cc|cc|cc|c|c@{}}
\toprule
\multirow{2}{*}{\textbf{Method}} & \multirow{2}{*}{\textbf{\hspace{2pt}Params $\downarrow$\hspace{2pt}}} & \multirow{2}{*}{\textbf{\hspace{2pt}Flops $\downarrow$\hspace{2pt}}} & \multicolumn{2}{c|}{\textbf{Skin}} & \multicolumn{2}{c|}{\textbf{Polyp}} & \multicolumn{2}{c|}{\textbf{Fundus}} & \multicolumn{2}{c|}{\textbf{Neoplasm}} & \textbf{Cell}    & \textbf{All} \\ \cmidrule(l){4-13} 
                                 &                                                          &                                                          & \textbf{ISIC17}  & \textbf{ISIC18} & \textbf{ETIS}   & \textbf{ColonDB}  & \textbf{DRIVE}    & \textbf{FIVES}   & \textbf{BUSI}   & \textbf{ThyroidXL}   & \textbf{CellSeg} & \textbf{Avg} \\ \midrule
\textbf{U-Net \cite{u-net}}                    & 34.53 M                                                    & 65.53 G                                                    & 83.07            & 86.67           & 76.85           & 83.95             & 71.20             & 75.77            & 74.04           & 71.16                & 71.52            & 77.14        \\
\textbf{AttnUNet \cite{attentionu-net}}                & 34.88 M                                                    & 66.64 G                                                    & 83.66            & 87.05           & 76.84           & 86.46             & 71.68             & 75.99            & 74.48           & 72.50                & 72.64            & 77.92        \\
\textbf{DeepLabv3+ \cite{deeplabv3+}}              & 39.76 M                                                    & 14.92 G                                                    & 83.84            & 88.64           & 90.73           & 91.92             & 69.59             & 75.12            & 76.81           & 73.46                & 71.90            & 80.22        \\
\textbf{UNet++ \cite{unet++}}                  & \underline{09.16 M}                                                    & 34.65 G                                                    & 82.98            & 87.46           & 77.40            & 87.88             & 72.94             & \underline{85.74}            & 74.46           & 83.94                & 78.30            & 81.23        \\
\textbf{nnU-Net \cite{nnU-Net}}                  & 31.29 M                                                      & 55.26 G                                                      & 83.23            & 88.53           & 80.13           & 91.63             & 75.43             & 76.10            & 76.46           & 86.08                & 83.53            & 82.34        \\
\textbf{PraNet \cite{pranet}}                  & 32.55 M                                                    & 06.93 G                                                    & 83.03            & 88.56           & 83.84           & 89.16             & 75.21             & 84.57            & 75.14           & 85.51                & 79.07            & 82.68        \\
\textbf{TransUNet \cite{transunet}}               & 105.32 M                                                   & 38.52 G                                                    & 85.00            & 89.16           & 87.79           & 91.63             & 74.98             & 83.54            & 78.30           & 85.77                & 79.08            & 83.92        \\
\textbf{Swin-Unet \cite{swin-unet}}                & 27.17 M                                                    & 06.20 G                                                    & 83.97            & 89.26           & 85.10            & 89.27             & 74.93             & 84.17            & 77.38           & 85.80                & 78.84            & 83.19        \\
\textbf{UCTransNet \cite{uctransnet}}              & 65.60 M                                                    & 56.70 G                                                    & 83.27            & 89.18           & 87.35           & 91.65             & 75.42             & 84.74            & 79.53           & 85.82                & 79.33            & 84.03        \\
\textbf{UNeXt \cite{unext}}                   & \textbf{1.470 M}                                                    & \textbf{0.570 G}                                                     & 82.74            & 87.78           & 74.03           & 83.84             & 74.77             & 76.60            & 74.71           & 84.46                & 75.71            & 79.40        \\
\textbf{VM-UNet \cite{vm-unet}}                 & 27.43 M                                                    & \underline{04.12 G}                                                    & \underline{85.99}            & 87.05           & 85.52           & 88.71             & 73.25             & 83.51            & 74.69           & 78.31                & 74.94            & 81.33        \\
\textbf{Swin-UMamba \cite{swin-umamba}}             & 60.00 M                                                    & 68.00 G                                                    & 83.40            & 87.62           & 86.63           & 87.97             & 73.32             & 82.66            & 73.38           & 84.96                & 75.56            & 81.72        \\
\textbf{EMCAD \cite{emcad}}                   & 26.76 M                                                    & 05.60 G                                                    & 85.95            & 90.96           & \underline{92.29}           & \underline{92.31}             & 77.15             & 82.51            & \underline{80.25}           & 83.33                & 79.13            & 84.87        \\
\textbf{MCADS \cite{mcads-decoder}}                   & 50.90 M                                                    & 61.89 G                                                    & 84.14            & \underline{91.01}           & 92.24           & 91.37             & \underline{78.42}             & 76.05            & 80.03           & \underline{86.33}                & \textbf{86.68}            & \underline{85.14}        \\ \midrule
\textbf{Ours}                    & 30.60 M                                                    & 05.00 G                                                    & \textbf{86.61}            & \textbf{91.56}           & \textbf{93.47}           & \textbf{93.27}             & \textbf{81.63}             & \textbf{87.50}            & \textbf{83.47}           & \textbf{88.02}                & \underline{86.61}            & \textbf{88.01}        \\ \midrule
\textbf{AutoSam \cite{autosam}*}                 & 41.56 M                                                    & 25.11 G                                                    & -                & -               & 79.70           & 83.00             & -                 & -                & -               & -                    & -                & -            \\
\textbf{Medical SAM3 \cite{medicalsam3}*}            & 840.0 M                                                    & -                                                        & -                & -           & 86.10           & -                 & 55.80             & -                & -               & -                    & -                & -            \\ \bottomrule
\end{tabular}%
}
\end{table*}

\begin{table}[]
\centering
\caption{Performance comparison with SOTA methods on the Synapse multi-organ dataset. Overall Dice, IoU, and HD95 are reported together with per class Dice scores. All methods were reproduced and averaged over five runs, with fine tuning applied to SOTA models for fair comparison. Results marked with * are reported from the papers. “–” indicates unavailable results.}
\label{tab:tab2}
\resizebox{1\columnwidth}{!}{%
\begin{tabular}{@{}l|ccc|cccccccc@{}}
\toprule
\textbf{Method}           & \textbf{Dice $\uparrow$} & \textbf{IoU $\uparrow$} & \textbf{HD95 $\downarrow$} & \textbf{Aorta} & \textbf{GB} & \textbf{KL} & \textbf{KR} & \textbf{Liver} & \textbf{PC} & \textbf{SP} & \textbf{SM} \\ \midrule
\textbf{U-Net \cite{u-net}}             & 70.11         & 59.39        & 44.69         & 84.00          & 56.70       & 72.41       & 62.64       & 86.98          & 48.73       & 81.48       & 67.96       \\
\textbf{AttnUNet \cite{attentionu-net}}         & 71.70         & 68.09        & 26.01         & 84.04          & 66.42       & 57.26       & 84.53       & 81.28          & 73.87       & 66.06       & 60.17       \\
\textbf{UNet++ \cite{unet++}}           & 72.39         & 68.82        & 25.61         & 83.65          & 67.66       & 57.26       & 84.53       & 81.34          & 73.87       & 68.97       & 61.85       \\
\textbf{nnU-Net \cite{nnU-Net}}           & 75.33         & 71.47        & 19.34         & 77.06          & 73.27       & 76.34       & 84.53       & 79.98          & 73.34       & 77.62       & 60.52       \\
\textbf{PraNetV2 \cite{pranet2}}         & 83.75         & 74.81        & 17.77         & 88.69          & 72.79       & 85.41       & 82.91       & \underline{95.82}          & 68.47       & \underline{93.09}       & \underline{85.85}       \\
\textbf{TransUNet \cite{transunet}}        & 77.61         & 67.32        & 26.90         & 86.56          & 60.43       & 80.54       & 78.53       & 94.33          & 58.47       & 87.06       & 75.00       \\
\textbf{Swin-Unet \cite{swin-unet}}         & 77.58         & 66.88        & 27.32         & 81.76          & 65.95       & 82.32       & 79.22       & 93.73          & 53.81       & 88.04       & 75.79       \\
\textbf{UCTransNet \cite{uctransnet}}       & 79.08         & 75.41        & 15.59         & 83.06          & 81.35       & 77.24       & 78.23       & 85.76          & 74.77       & 81.89       & 70.31       \\
\textbf{UNETR \cite{unetr}*}            & 78.35         & -            & 18.59         & 89.80          & 56.30       & 85.60       & 84.52       & 94.57          & 60.47       & 85.00       & 70.46       \\
\textbf{MISSFormer \cite{missformer}*}       & 81.96         & -            & 18.20         & 86.99          & 68.65       & 85.21       & 82.00       & 94.41          & 65.67       & 91.92       & 80.81       \\
\textbf{U-Mamba \cite{u-mamba}}          & 78.63         & 74.87        & 16.19         & 83.77          & 78.70       & 79.40       & 82.37       & 83.86          & 74.78       & 79.77       & 66.41       \\
\textbf{VM-UNet \cite{vm-unet}}          & 73.39         & 71.61        & 27.97         & 63.57          & 72.62       & 77.98       & \textbf{92.59 }      & 79.44          & 70.80       & 55.58       & 74.55       \\
\textbf{EMCAD \cite{emcad}}            & 83.63         & 74.65        & 15.68         & 88.14          & 68.87       & \underline{88.08}       & 84.10       & 95.26          & 68.51       & 92.17       & 83.92       \\
\textbf{MCADS \cite{mcads-decoder}}            & 85.03         & \underline{81.71}        & \textbf{11.11}         & 90.81          & \underline{86.07}       & 86.77       & 83.24       & 87.66          & \textbf{83.55}       & 85.74       & 76.38       \\ \midrule
\textbf{Ours}             & \textbf{87.00}$\pm0.2$         & \textbf{83.77}        & \underline{14.39}         & \textbf{92.28}          & \textbf{90.31}       & \textbf{89.72}       & \underline{87.21}       & 91.02          & \underline{82.08}       & 86.91       & 76.51       \\ \midrule
\textbf{Self -Prompt SAM \cite{self-promptsam}*} & \underline{86.74}         & -            & -             & \underline{91.99}          & 69.95       & 85.65       & 85.40       & \textbf{97.39}          & 79.18       & \textbf{94.38}       & \textbf{89.94}       \\ \bottomrule
\end{tabular}%
}
\end{table}

\section{Experiments}
\subsection{Datasets and Evaluation Metrics}
We evaluated the proposed method on 11 publicly available medical image segmentation datasets that have also been benchmarked in recent SOTA studies, including EMCAD (CVPR 2024) \cite{emcad}, Swin-UMamba (MICCAI 2024) \cite{swin-umamba}, ThyroidXL (MICCAI 2025) \cite{thyroidxl}, and Medical-SAM3 \cite{medicalsam3} (2026). The datasets span diverse organs and imaging modalities, including ISIC17 \cite{isic17}, ISIC18 \cite{isic18}, ETIS \cite{etis}, ColonDB \cite{etis}, DRIVE \cite{drive}, FIVES \cite{fives}, BUSI \cite{busi}, ThyroidXL \cite{thyroidxl}, CellSeg \cite{cellseg}, Synapse \cite{synapse}, and ACDC \cite{acdc}, covering dermoscopy, endoscopy, fundus imaging, ultrasound, microscopy, CT, and MRI. Performance was assessed using Dice, IoU, and HD95 \cite{miseval}. Additional dataset and metric details are provided in the supplementary material.

\subsection{Implementation details}
We implemented the proposed network in PyTorch 2.7 and conducted all experiments on a single NVIDIA RTX 3090 GPU with 24 GB of memory. The model was optimized using AdamW with a learning rate of 1e-4. Training was performed for over 300 epochs with a batch size of 16, and the best model was selected based on the validation Dice score. All input images were resized to 224 by 224, and online data augmentation including random rotation, horizontal and vertical flipping, and random cropping was applied. The network was trained using a BCE loss. For the DRIVE and FIVES datasets, we generated 256 by 256 overlapping patches with a stride of 128 for training. For CellSeg, we generated 384 by 384 overlapping patches with a stride of 192 for training. During testing, similar overlapping patches were extracted, predictions were obtained for each patch, and the full resolution segmentation maps were reconstructed for evaluation. To ensure a fair comparison, all competing methods were reproduced using their publicly available implementations, and the results were averaged over five independent runs.

\subsection{Results}
We compare our method with representative CNN, transformer, Mamba, SAM, and decoder centric models on 2D binary and multi-class benchmarks. Across the 9 binary-class segmentation datasets in Table \ref{tab:tab1}, our method consistently outperforms all baselines on skin, polyp, fundus, neoplasm, and cell tasks. Conventional CNNs remain stable but are constrained by limited global modeling, while nnU-Net improves performance through optimization without closing the performance gap. Transformer models strengthen global reasoning, with UCTransNet achieving competitive results at the cost of higher architectural complexity. Mamba variants model long-range dependencies efficiently yet yield only marginal or unstable gains. Decoder centric approaches, especially EMCAD and MCADS, form the strongest baselines, underscoring the importance of feature fusion. However, our results demonstrate that structured context-aware skip gating yields superior performance without relying on larger or heavier designs. SAM based foundation models underperform on domain specific medical data, highlighting the necessity of task tailored architectures. On the Synapse multi-class segmentation dataset in Table \ref{tab:tab2}, our method achieves the highest overall Dice and IoU while remaining competitive in boundary accuracy. CNN baselines struggle with small and complex organs, transformers improve structural coherence but exhibit class level variability, and Mamba models do not consistently minimize boundary errors. Although MCADS demonstrates strong boundary performance, our method maintains a better overall balance between accuracy and structural consistency. On the ACDC multi-class segmentation dataset in Table \ref{tab:tab3}, our method achieves the best Dice and the lowest HD95, clearly surpassing CNN, transformer, and Mamba models. The substantial reduction in HD95 reflects sharper boundary delineation across RV, Myo, and LV, confirming that context-aware gated fusion enhances both anatomical coherence and fine structural precision. Overall, simply increasing encoder scale or global modeling capacity is insufficient. Effective decoder design is decisive, as reflected by consistently superior performance over strong baselines.

In terms of \textbf{computational cost}, our method maintains a strong efficiency profile, as evident in Table \ref{tab:tab1}, while delivering superior segmentation accuracy. Compared with SOTA decoder centric approaches EMCAD and MCADS, our design achieves higher performance with fewer parameters and comparable or lower FLOPs.

\begin{figure}
\centerline{\includegraphics[width=0.9\textwidth]{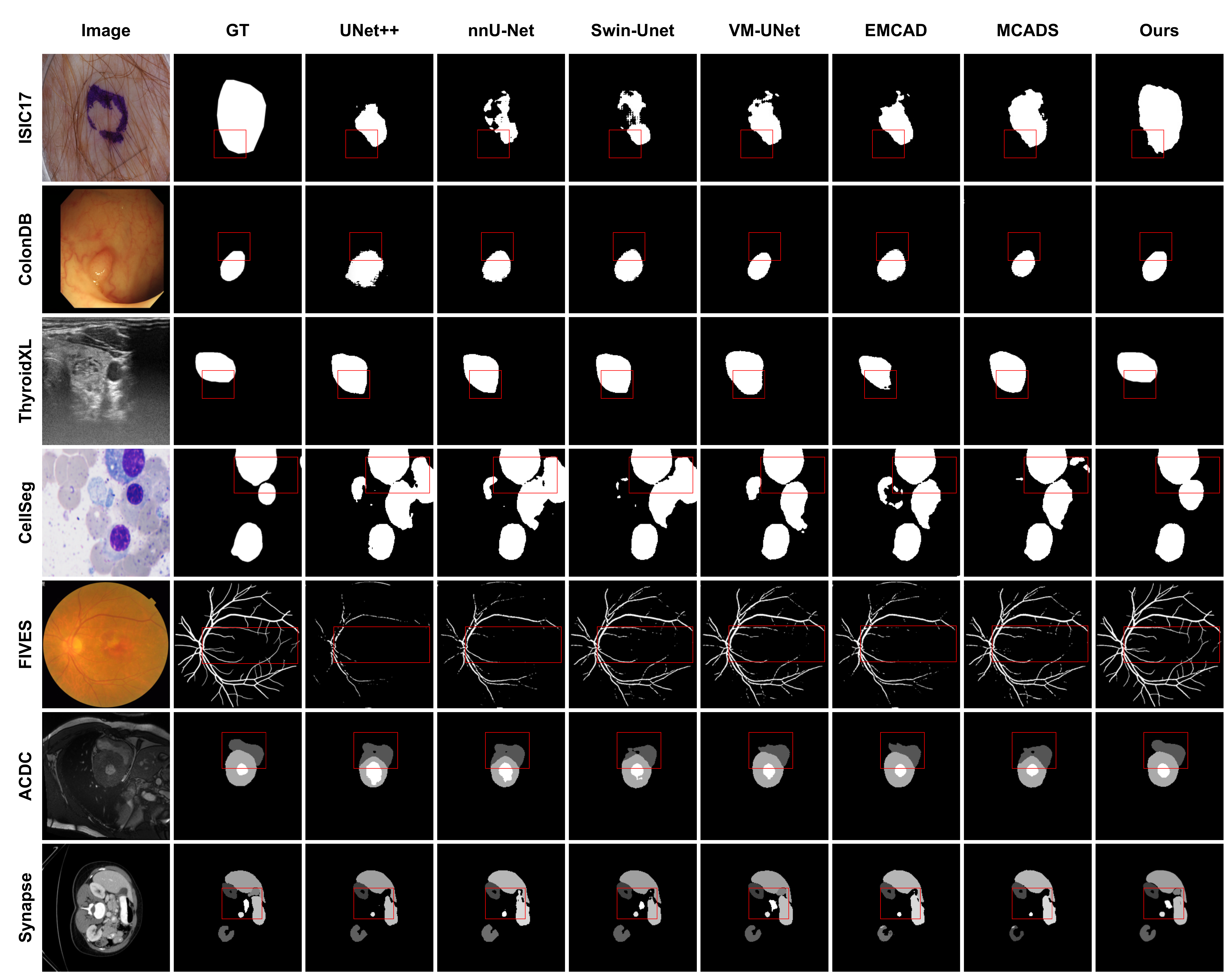}}
\caption{Qualitative Results Comparison. Red rectangles highlight incorrect segmentation regions.}
\label{fig:qual}
\end{figure}

\textbf{Qualitative results} are provided in Fig. \ref{fig:qual}, showing superior segmentation performance across all tasks. For vessel segmentation in fundus images, our method accurately captures nearly all vessels, whereas other methods struggle to recover complex tree like structures and often miss many branches. In polyp, skin, thyroid, and cell segmentation, our approach better preserves region shapes while avoiding over segmentation, a common issue observed in several CNN and Mamba-based methods. For multi-class segmentation on the ACDC and Synapse datasets, most methods produce reasonable results; however, CNN based models often miss regions, while EMCAD and MCADS occasionally fail to detect certain classes.

\section{Ablation Studies}
\label{sec:ablstd}
In this section, we conduct ablation studies to analyze the key architectural components and design choices of the proposed decoder, isolating their individual and combined contributions to segmentation performance through systematic empirical evaluation. All experiments are performed on the Synapse multi-organ dataset for multi-class segmentation and the CellSeg dataset for binary segmentation to ensure reliable evaluation across settings.

\subsection{Component-Level Analysis of Decoder Architectural Design Choices}
\label{sec:ablcomponent}
\begin{figure}
\centerline{\includegraphics[width=0.9\textwidth]{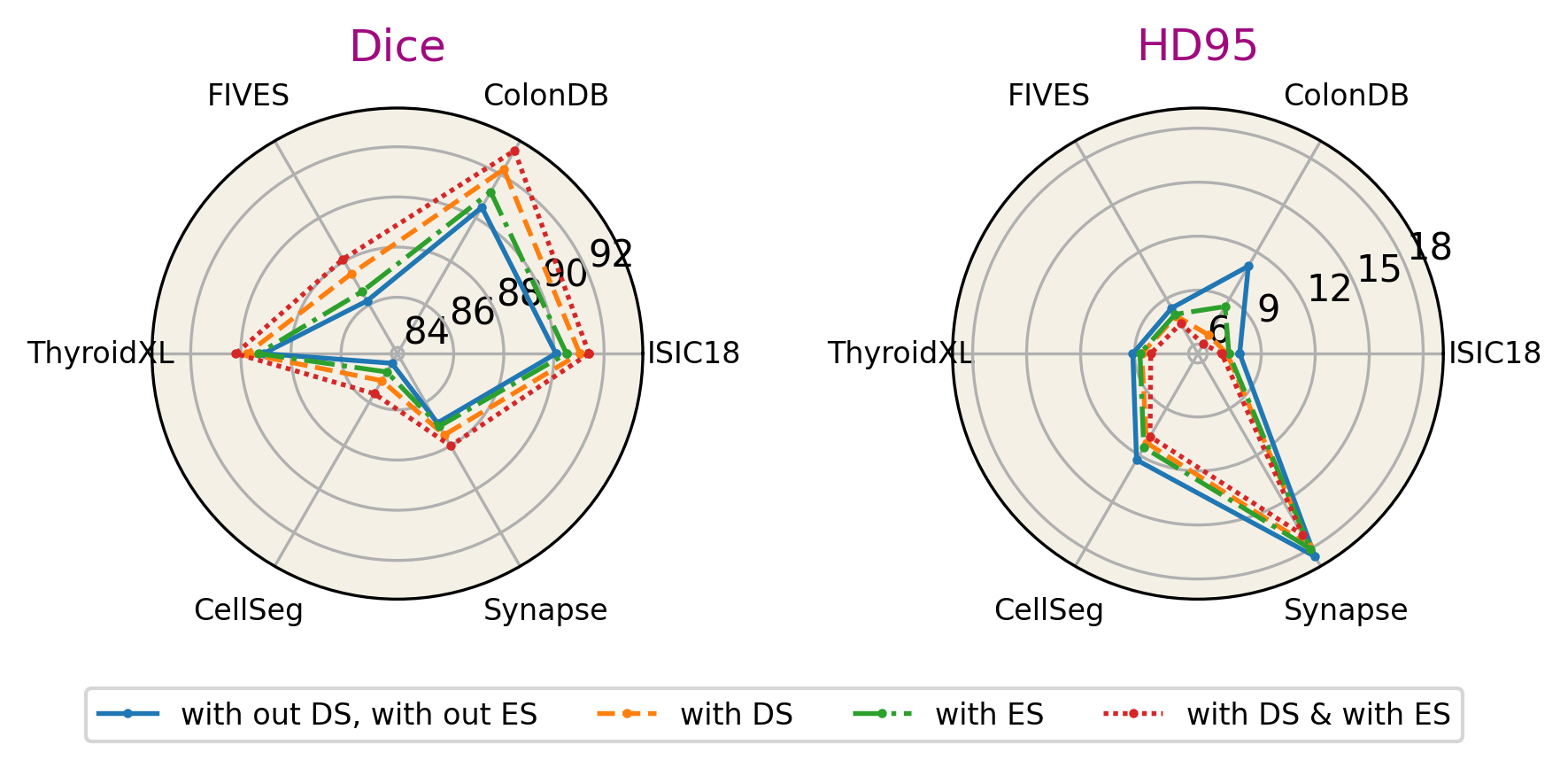}}
\caption{Radar plots showing the effect of Deep Supervision (DS) and Edge Supervision (ES) across six segmentation benchmarks. In the Dice plot ($\uparrow$), performance improves as values move toward the outer rings. In the HD95 plot ($\downarrow$), lower values are better, so profiles closer to the center indicate more accurate boundaries. Compared with using either supervision alone or neither, enabling both DS and ES consistently achieves the best overall performance across all datasets.}
\label{fig:abl4}
\end{figure}

We conduct a component-level ablation to analyze the individual and cumulative contributions of each decoder module, as summarized in Table~\ref{tab:abl1}.  The baseline employs a PVTv2-B2 encoder with a plain U-Net style decoder, revealing the limitations of naive upsampling and direct skip fusion. Adding BT improves performance by injecting global context at the deepest stage. Enabling CA without RA further enhances accuracy through multi-scale aggregation, though gains remain limited due to the lack of explicit global modulation. Incorporating RA within CA yields a larger improvement, highlighting the importance of residual global stabilization. Adding RB strengthens reconstruction, and integrating SCA-Gate delivers a clear boost, showing that competitive and structured skip regulation is more effective than direct concatenation.
Finally, we evaluate Stage 0, which introduces an additional refinement pathway from the raw input. It applies RB and SCA-Gate after the Stage 1 2x upsampling. Although it yields slight improvements, the gains are marginal and inconsistent across tasks, and the total computational cost increases to 8.317 GFLOPs. Given this unfavorable trade off, Stage 0 is not included in the final model.
Fig. \ref{fig:abl4} shows a radar plot of Deep Supervision (DS) and Edge Supervision (ES) settings across six benchmarks. In the Dice plot, enabling both DS and ES covers the largest area, showing the best segmentation accuracy. In the HD95 plot, the same setting covers the smallest area, indicating lower boundary error and better contour accuracy.
Overall, performance improves consistently with BT, CA with RA, RB, and SCA-Gate. Accordingly, the final decoder configuration is directly guided by the empirical evidence, where each retained component demonstrates consistent and complementary gains.\\
\begin{wraptable}[18]{r}{0.5\textwidth}
\centering
\caption{Performance comparison with SOTA methods on the ACDC dataset. Overall Dice, IoU, and HD95 are reported together with per class Dice scores. All methods were reproduced and averaged over five runs, with fine tuning applied to SOTA models for fair comparison.}
\label{tab:tab3}
\resizebox{0.5\columnwidth}{!}{%
\begin{tabular}{@{}l|ccc|ccc@{}}
\toprule
\textbf{Method}     & \textbf{Dice $\uparrow$} & \textbf{IoU $\uparrow$} & \textbf{HD95 $\downarrow$} & \textbf{RV} & \textbf{Myo} & \textbf{LV} \\ \midrule
\textbf{U-Net \cite{u-net}}       & 81.56         & 73.41        & 6.9854        & 76.99       & 80.28        & 87.43       \\
\textbf{AttnUNet \cite{attentionu-net}}   & 82.37         & 73.94        & 6.1684        & 78.13       & 81.08        & 87.89       \\
\textbf{UNet++ \cite{unet++}}     & 81.97         & 73.92        & 6.4724        & 77.74       & 80.73        & 87.44       \\
\textbf{nnU-Net \cite{nnU-Net}}     & 82.66         & 74.27        & 6.1663        & 79.00          & 81.01        & 87.97       \\
\textbf{PraNetV2 \cite{pranet2}}   & 83.74         & 76.13        & 6.3719        & 79.61       & 83.10         & 88.51       \\
\textbf{TransUNet \cite{transunet}}  & 83.07         & 74.85        & 5.7578        & 79.16       & 81.65        & 88.41       \\
\textbf{Swin-Unet \cite{swin-unet}}   & 82.61         & 74.59        & 6.1244        & 78.94       & 80.17        & 88.73       \\
\textbf{UCTransNet \cite{uctransnet}} & 84.89         & 77.57        & 5.6995        & 80.94       & 84.11        & \underline{89.62}       \\
\textbf{U-Mamba \cite{u-mamba}}    & 84.18         & 76.47        & 5.8501        & 80.90        & 83.24        & 88.40        \\
\textbf{VM-UNet \cite{vm-unet}}    & 81.02         & 72.74        & 7.0025        & 76.75       & 79.40         & 86.90        \\
\textbf{EMCAD \cite{emcad}}      & \underline{85.07}         & \underline{77.73}        & \underline{5.2472}        & \underline{81.58}       & \underline{84.23}        & 89.42       \\
\textbf{MCADS \cite{mcads-decoder}}      & 84.51         & 76.92        & 5.5595        & 81.16       & 83.27        & 89.09       \\ \midrule
\textbf{Ours}       & \textbf{87.54}$\pm0.3$         & \textbf{80.96 }       & \textbf{4.4057}        & \textbf{85.27}       & \textbf{86.23}        & \textbf{91.11}       \\ \bottomrule
\end{tabular}%
}
\end{wraptable}
\subsection{Comparison with Baseline Attention Mechanisms for Skip Connection}
\label{sec:ablgate}
To validate SCA-Gate, we compare it with representative skip attention mechanisms in Table \ref{tab:ab2}, where the baseline is our full model without any attention gate. The baseline already performs strongly, confirming that gains are not solely due to the backbone as evident from Table \ref{tab:abl1}. Attention U-Net Gate yields a moderate improvement but remains limited by its simple additive gating formulation. Attention U-Net Gate with ECA provides a slight additional gain through enhanced channel sensitivity, yet still lacks explicit multi-scale spatial competition. Attention U-Net Gate with our ECA-MSP provides additional gain, though it remains within a conventional gating framework. LGAG from EMCAD offers competitive performance by enlarging local context through grouped convolution based gating, but primarily emphasizes spatial refinement. RLAB from MCADS Decoder delivers stable gains via residual based refinement, yet does not explicitly model competitive encoder-decoder alignment. In contrast, SCA-Gate achieves the highest overall performance by jointly modeling spatial competition and channel aware contextual modulation for selective and semantically aligned skip transmission.

\subsection{Backbone Variants and Resolution Analysis}
\label{sec:ablback}
Table \ref{tab:abl3} compares pretrained PyTorch timm encoders under different input resolutions, where higher resolution improves segmentation accuracy but significantly increases computational cost, As expected, performance scales with encoder strength while remaining competitive even with lighter backbones, underscoring that our claim is compatibility rather than identical performance across encoder families. Convnext shows lower accuracy, maxvit achieves higher accuracy, and swin again attains relatively lower performance, while all remain heavier than the selected backbone. At 512x512 resolution, pvt\_v2 improves accuracy over most 224 settings, and maxvit achieves the highest performance overall, but both require substantially greater computational resources. In comparison, our selected pvt\_v2\_b2 with 224x224 input provides a more practical balance between performance and efficiency, as higher resolution variants offer gains at the cost of substantial computational overhead.

\begin{table}[]
\centering
\caption{Ablation study of decoder components. Average Dice scores $\uparrow$ are reported.}
\label{tab:abl1}
\resizebox{1\columnwidth}{!}{%
\begin{tabular}{@{}cccccc|ccc@{}}
\toprule
\multicolumn{1}{c|}{\textbf{Bottleneck}} & \multicolumn{1}{c|}{\textbf{Context Aggregator w RA}} & \multicolumn{1}{c|}{\textbf{Context Aggregator w/o RA}} & \multicolumn{1}{c|}{\textbf{Refinement Block}} & \multicolumn{1}{c|}{\textbf{SCA-Gate}} & \textbf{Stage 0} &  & \textbf{Synapse} & \textbf{CellSeg} \\ \midrule
\ding{55}                                & \ding{55}                                             & \ding{55}                                               & \ding{55}                                      & \ding{55}                              & \ding{55}        &  & 73.91            & 81.07            \\
\checkmark                               & \ding{55}                                             & \ding{55}                                               & \ding{55}                                      & \ding{55}                              & \ding{55}        &  & 75.53            & 82.80             \\
\checkmark                               & \ding{55}                                             & \checkmark                                              & \ding{55}                                      & \ding{55}                              & \ding{55}        &  & 79.38            & 82.60             \\
\ding{55}                                & \checkmark                                            & \ding{55}                                               & \ding{55}                                      & \ding{55}                              & \ding{55}        &  & 81.03            & 82.47            \\
\checkmark                               & \checkmark                                            & \ding{55}                                               & \ding{55}                                      & \ding{55}                              & \ding{55}        &  & 83.57            & 84.28            \\
\checkmark                               & \checkmark                                            & \ding{55}                                               & \checkmark                                     & \ding{55}                              & \ding{55}        &  & 85.19            & 84.62            \\
\checkmark                               & \checkmark                                            & \ding{55}                                               & \checkmark                                     & \checkmark                             & \ding{55}        &  & \underline{87.00}            & \underline{86.61}            \\
\checkmark                               & \checkmark                                            & \ding{55}                                               & \checkmark                                     & \checkmark                             & \checkmark       &  & \textbf{87.21}            & \textbf{86.63}            \\ \bottomrule
\end{tabular}%
}
\end{table}

\begin{table}[]
\centering
\caption{Comparative Analysis of Skip Attention Modules. Average Dice scores $\uparrow$ are reported.}
\label{tab:ab2}
\resizebox{1\columnwidth}{!}{
\begin{tabular}{@{}lllllll|llllllclllllllcllllll|llllcllllllc@{}}
\toprule
\textbf{Skip Connection Attention}       &  &  &  &  &  &  &  &  &  &  &  &  & \textbf{Params}  &  &  &  &  &  &  &  & \textbf{Flops}   &  &  &  &  &  &  &  &  &  &  & \textbf{Synapse} &  &  &  &  &  &  & \textbf{CellSeg} \\ \midrule
Attention U-Net - Gate           &  &  &  &  &  &  &  &  &  &  &  &  & \underline{29.54} M &  &  &  &  &  &  &  & \underline{4.626} G &  &  &  &  &  &  &  &  &  &  & 85.20   &  &  &  &  &  &  & 83.19   \\
Attention U-Net - Gate + ECA     &  &  &  &  &  &  &  &  &  &  &  &  & 29.58 M &  &  &  &  &  &  &  & \textbf{4.625} G &  &  &  &  &  &  &  &  &  &  & 85.38   &  &  &  &  &  &  & 80.77   \\
Attention U-Net - Gate + ECA-MSP &  &  &  &  &  &  &  &  &  &  &  &  & \textbf{29.45} M &  &  &  &  &  &  &  & 4.626 G &  &  &  &  &  &  &  &  &  &  & \underline{85.46}   &  &  &  &  &  &  & \underline{84.65}   \\
LGAG (EMCAD)                    &  &  &  &  &  &  &  &  &  &  &  &  & 30.94 M &  &  &  &  &  &  &  & 4.991 G &  &  &  &  &  &  &  &  &  &  & 84.51   &  &  &  &  &  &  & 81.99   \\
RLAB (MCADS-Decoder)            &  &  &  &  &  &  &  &  &  &  &  &  & 30.96 M &  &  &  &  &  &  &  & 5.861 G &  &  &  &  &  &  &  &  &  &  & 85.17   &  &  &  &  &  &  & 80.37   \\
SCA-Gate (Ours)                 &  &  &  &  &  &  &  &  &  &  &  &  & 30.60 M &  &  &  &  &  &  &  & 5.001 G &  &  &  &  &  &  &  &  &  &  & \textbf{87.00}   &  &  &  &  &  &  & \textbf{86.61}   \\ \bottomrule
\end{tabular}%
}
\end{table}

\begin{table}[]
\centering
\caption{Comparison of Different Encoder Backbones. Average Dice scores $\uparrow$ are reported.}
\label{tab:abl3}
\resizebox{1\columnwidth}{!}{%
\begin{tabular}{@{}lll|llcll|llcllllcll|llcllllc@{}}
\toprule
\textbf{Encoder}                          &  &  &  &  & \textbf{Input Size} &  &  &  &  & \textbf{Params} &  &  &  &  & \textbf{Flops} &  &  &  &  & \textbf{Synapse} &  &  &  &  & \textbf{CellSeg} \\ \midrule
\textbf{convnext\_base}                   &  &  &  &  & 224x224             &  &  &  &  & 93.72 M           &  &  &  &  & \textbf{16.538} G         &  &  &  &  & 86.79            &  &  &  &  & 84.13            \\
\textbf{maxvit\_base\_tf\_224}            &  &  &  &  & 224x224             &  &  &  &  & \underline{84.19} M           &  &  &  &  & 50.317 G         &  &  &  &  & 87.57            &  &  &  &  & \underline{86.79}            \\
\textbf{swin\_base\_patch4\_window7\_224} &  &  &  &  & 224x224             &  &  &  &  & 92.89 M           &  &  &  &  & \underline{16.622} G         &  &  &  &  & 86.52            &  &  &  &  & 85.50            \\
\textbf{pvt\_v2\_b2}                      &  &  &  &  & 512x512             &  &  &  &  & \textbf{30.61} M           &  &  &  &  & 67.495 G         &  &  &  &  & \underline{87.99}            &  &  &  &  & 86.75            \\
\textbf{maxvit\_base\_tf\_512}            &  &  &  &  & 512x512             &  &  &  &  & 84.57 M           &  &  &  &  & 179.784 G        &  &  &  &  & \textbf{88.95}            &  &  &  &  & \textbf{87.94}            \\ \bottomrule
\end{tabular}%
}
\end{table}

\section{Conclusion}
In this work, we revisit medical image segmentation from a decoder-centric perspective. We introduced a context-aware gated decoding framework for medical image segmentation that integrates global context modeling and adaptive skip fusion within a unified encoder-decoder design. By refining multi-scale features in a shared space, the method improves semantic consistency and boundary accuracy while maintaining strong computational efficiency compared to recent SOTA approaches, achieving consistent gains across diverse benchmarks. While this work focuses on improving decoder accuracy and efficiency under standardized 2D evaluation protocols, extending the framework to OOD robustness, 3D segmentation, and comparisons with large scale foundation models remains an important direction for future research.

\section{Acknowledgments}
This work was supported by the National Research Foundation of Korea(NRF) grant funded by the Korea government(MSIT)(RS-2025-00573160), the Institute of Information \& Communications Technology Planning \& Evaluation(IITP)-Innovative Human Resource Development for Local Intellectualization program grant funded by the Korea government(MSIT)(IITP-2026-RS-2020-II201489), and the “Advanced GPU Utilization Support Program” funded by the Government of the Republic of Korea (Ministry of Science and ICT).


%
%
\bibliographystyle{splncs04}
\bibliography{main}

\newpage



\section*{Supplementary material}

# MedCAGD: Context-Aware Gated Decoder for Efficient Medical Image Segmentation
## Supplementary Material

Saad Wazir[1], Patrick Dominique Vibild[2], Dinh Phu Tran[1], Seongah Kim[1], and Daeyoung Kim[1]

[1] School of Computing, Korea Advanced Institute of Science and Technology (KAIST), Daejeon, Republic of Korea
{saad.wazir,phutx2000,kimsa0322,kimd}@kaist.ac.kr
[2] Department of Energy, Aalborg University, Aalborg, Denmark
padovi@energy.aau.dk

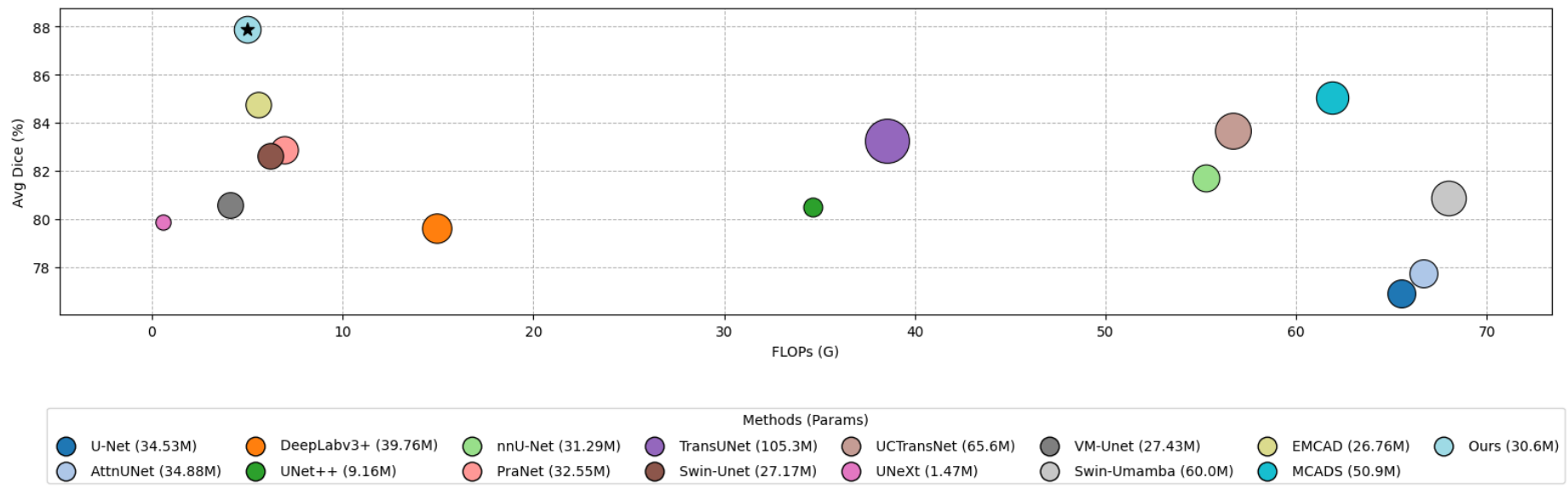


**Fig. 4:** Efficiency comparison of segmentation methods in terms of computational cost and performance. The scatter plot shows the relationship between FLOPs (G) and average Dice score (%) across different segmentation models. Each marker represents a method, where the x-axis indicates computational complexity (FLOPs) and the y-axis denotes segmentation accuracy (Avg Dice). The marker size is proportional to the number of parameters (M), reflecting model capacity, while different colors distinguish individual methods. Our method is highlighted with a star marker for clear identification. A desirable model appears toward the upper-left region, indicating higher segmentation accuracy with lower computational cost.

## 7 Experiment Details

This section extends Section 4 of the original paper by detailing the datasets and evaluation metrics, followed by additional experimental results.

### 7.1 Dataset Details

To evaluate the performance of our MedCAGD decoder, we conduct experiments on 11 datasets covering eight medical image segmentation tasks, as described below.

i. **Skin lesion segmentation:** These datasets consist of dermoscopic color images of the skin surface and are designed for melanoma detection. They provide data for binary-class semantic segmentation of skin lesion regions. For the ISIC18 [3] dataset, following the EMCAD [10] experimental protocol, we use 2,594 images with an 80:10:10 split for training, validation, and testing. For ISIC17 [4], following EMCAD [10], we use the official split consisting of 2,000 training, 150 validation, and 600 testing images.

ii. **Polyp segmentation:** These datasets contain colonoscopy images collected from multiple imaging centers and clinics, providing diversity in imaging conditions as well as variations in polyp size, shape, and appearance. For ColonDB (379 images) and ETIS (196 images) [12], following the EMCAD [10] protocol, we use an 80:10:10 split for training, validation, and testing.

iii. **Fundus vessel segmentation:** These datasets consist of color retinal fundus photographs and are designed for binary-class semantic segmentation of retinal vessels. The images include both healthy subjects and patients with diseases such as diabetic retinopathy (DR), age-related macular degeneration (AMD), and glaucoma. We follow the official dataset distributions. For the FIVES [6] dataset, 600 images are used for training and 200 images for testing. For DRIVE [11], 20 images are used for training and 20 images for testing. A validation set is created from the training data using a 90:10 split.

iv. **Breast cancer segmentation:** This dataset consists of B-mode breast ultrasound scans and is designed for binary-class semantic segmentation of breast lesions and tumors. For the BUSI [1] dataset, following EMCAD [10], we use 647 images (437 benign and 210 malignant) and apply an 80:10:10 split for training, validation, and testing.

v. **Thyroid segmentation:** Advancing thyroid nodule diagnosis (ThyroidXL) [5] is an expert-labeled and pathology-validated dataset of B-mode ultrasound images designed for binary-class semantic segmentation of thyroid nodule regions. We follow the official distribution, which contains 9,541 image-mask pairs for training and 2,094 pairs for testing. A validation set is created from the training set using a 90:10 split.

vi. **Cell segmentation:** This dataset contains high-resolution microscopy images acquired from multiple modalities, including Brightfield, Fluorescent, and Phase Contrast (PC), capturing diverse cellular structures and imaging conditions. We utilize the CellSeg [8] dataset for binary-class semantic segmentation of cells, capturing variations such as irregular shapes, elongated forms, and weak boundaries. We use the official split consisting of 1,000 training samples and 50 testing samples from biological imaging datasets for cell nuclei and structure segmentation. A validation set is created from the training data using a 90:10 split.

vii. **Abdominal organ segmentation:** This dataset consists of abdominal CT scans designed for multi-class semantic segmentation of multiple abdominal organs with complex anatomical annotations. For the Synapse multi-organ dataset [7], following EMCAD [10], we use 30 abdominal CT scans containing 3,779 axial contrast-enhanced slices. Each scan includes 85 to 198 slices of size $512 \times 512$. Following EMCAD [10], 18 scans (2,212 slices) are used for training and 12 scans for val-

idation. We segment eight abdominal organs: aorta, gallbladder (GB), left kidney (KL), right kidney (KR), liver, pancreas (PC), spleen (SP), and stomach (SM).

viii. **Cardiac organ segmentation:** This dataset contains cardiac cine magnetic resonance imaging (MRI) scans and supports multi-class semantic segmentation of anatomical cardiac structures across healthy hearts and various pathologies, including Dilated Cardiomyopathy and Myocardial Infarction. We use the ACDC [2] dataset and follow the official distribution. The dataset contains 100 cardiac MRI scans for training and 50 scans for testing, with three annotated organs: right ventricle (RV), myocardium (Myo), and left ventricle (LV). A validation set is created from the training data using a 90:10 split.

### 7.2 Evaluation Metrics

To evaluate state-of-the-art deep learning methods and our proposed work, we have used standard evaluation metrics [9]. All the evaluation metrics are reported by calculating each evaluation metric for each prediction and taking the average over all samples in the test set. We use the threshold of 0.5. Following evaluation metrics are used:

i. **Intersection over Union (IoU)** is defined as

$$\text{IoU} = \frac{TP}{TP + FP + FN} \tag{1}$$

Where TP are True Positives , FP are False Positives, and FN are False Negatives. Intersection = TP, and Union = TP + FP + FN.

ii. **Dice Coefficient (DSC)** is defined as

$$\text{DSC} = \frac{2 \times \text{TP}}{2 \times \text{TP} + \text{FP} + \text{FN}} \tag{2}$$

iii. **95th Percentile Hausdorff Distance (HD95)** measures the boundary discrepancy between predicted and ground truth segmentation's by computing the 95th percentile of nearest point distances between the two sets, reducing sensitivity to outliers. Lower HD95 values indicate better boundary alignment.
HD95 is defined as:

$$HD95 = \max\left(\text{perc}_{95}(\min_{b \in B} d(a,b)),\ \text{perc}_{95}(\min_{a \in A} d(b,a))\right) \tag{3}$$

where $d(a, b)$ is the Euclidean distance between points $a$ and $b$, and $\text{perc}_{95}$ denotes the 95th percentile of the computed distances.

### 7.3 Additional Experimental Results

The computational efficiency vs. segmentation accuracy graph is shown in Fig. 4, comparing SOTA methods with our method. The Dice scores from all 11 datasets are averaged, demonstrating that our method achieves the highest segmentation accuracy while

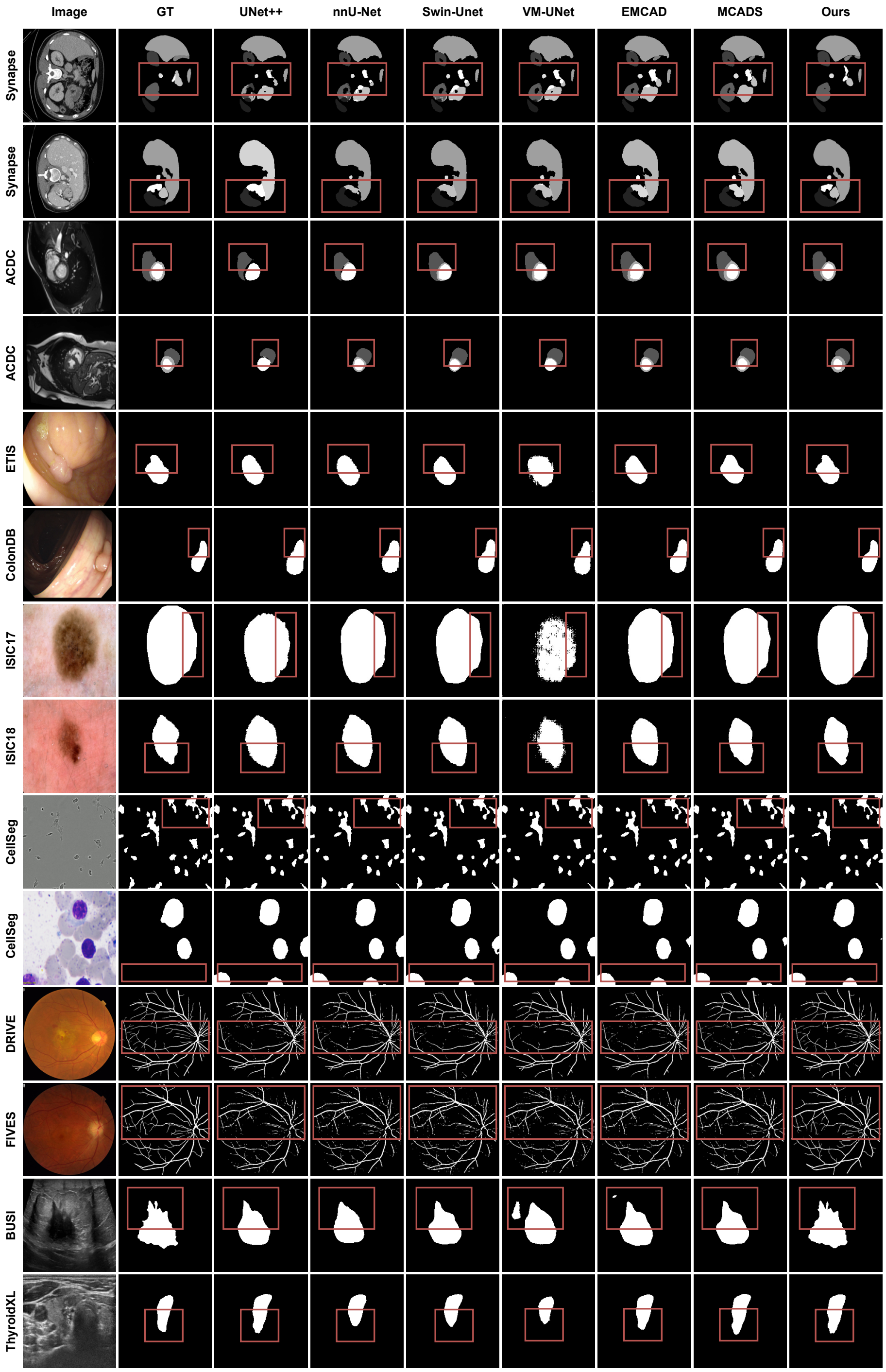


**Fig. 5:** Additional qualitative comparisons. Red rectangles indicate regions with incorrect segmentation

maintaining practically feasible and low resource consumption. Fig. 5 presents additional qualitative comparisons. The examples show standard and challenging cases. Red rectangles indicate regions with incorrect segmentation. To facilitate analysis, we discuss the qualitative results for each dataset below.

On the Synapse dataset, baseline methods frequently misclassify several classes, whereas our proposed method successfully predicts all classes with high accuracy. Both EMCAD and MCADS fail to properly separate objects in the more challenging classes, while traditional CNN-based approaches exhibit the weakest performance. In contrast, our method effectively isolates objects in difficult categories while completely avoiding over-segmentation.

On the ACDC dataset, UNet++ completely fails to predict one of the classes. Although other baseline methods struggle to maintain accurate boundary delineation, our method consistently produces clear and precise boundaries. Notably, even in relatively less complex cases, UNet++, nnU-Net, and VM-UNet fail to achieve optimal segmentation quality.

Across the skin and polyp datasets, our method demonstrates superior performance in challenging cases, particularly by achieving higher boundary accuracy and reducing missed predictions. In contrast, VM-UNet fails completely in certain instances. Swin-Unet and MCADS reduce the number of missed predictions, whereas EMCAD and other comparative methods tend to produce over-segmentation.

For cell segmentation, difficult cases often contain nearly identical cells with available ground truth (GT) segmentation masks. In such scenarios, all baseline methods exhibit noticeable over-segmentation. Our method effectively addresses this challenge by generating highly accurate cell predictions with clearly separated boundaries. MCADS also shows competitive performance compared to the other baselines.

On the fundus vessel segmentation datasets (FIVES and DRIVE), our method achieves superior segmentation accuracy by successfully outlining thin, connecting vessels with excellent boundary precision. In contrast, all other baseline methods struggle to correctly segment these delicate vascular structures.

For the BUSI dataset, our method produces the most reliable qualitative results with highly visible boundary accuracy. Although MCADS and EMCAD reduce over-segmentation, they struggle to preserve anatomical shapes and fail to segment more challenging regions. VM-UNet, on the other hand, exhibits severe over-segmentation.

On the ThyroidXL dataset, EMCAD and MCADS perform better than earlier baseline methods; however, VM-UNet suffers from frequent missed predictions. In comparison, our method maintains clear and highly accurate segmentation results, even in the most challenging apical regions that are typically difficult to segment.

Overall, VM-UNet shows the weakest performance across the datasets. It frequently misses predictions, leaves segmentation patches (clearly visible in the skin and polyp datasets), misclassifies regions in multi-class datasets, and consistently produces over-segmentation. UNet++, nnU-Net, and Swin-Unet perform reasonably well in simpler cases but often fail in challenging scenarios, where they tend to miss predictions and produce inaccurate boundaries. EMCAD and MCADS demonstrate relatively strong performance and generally avoid complete class omissions; however, they struggle to

maintain boundary precision and perform poorly when segmenting thin vascular structures.

In contrast, our proposed method achieves the best overall qualitative performance. It consistently predicts all classes, generates highly accurate boundaries, minimizes over-segmentation, and uniquely succeeds in segmenting thin vessels as well as other challenging anatomical regions.



\end{document}